\documentclass[letterpaper, 10 pt, conference]{ieeeconf}  % Comment this line out if you need a4paper

\IEEEoverridecommandlockouts                              % This command is only needed if 
\usepackage{times}
\usepackage{amsmath}
\usepackage{calc}
\usepackage[T1]{fontenc}
\usepackage{amssymb}
\usepackage{amsmath}
\usepackage{amstext}
\usepackage{graphicx,color}
\usepackage{booktabs,tabularx}
\usepackage{multirow}
\usepackage{url}
\usepackage{tabularx}
\graphicspath{{images/}}
\usepackage{caption}
\usepackage{subcaption}
\usepackage{tabularx} % in the preamble
\usepackage{wrapfig}
\usepackage{caption}

\makeatletter
\let\NAT@parse\undefined
\renewcommand{\@IEEEsectpunct}{ }% Modified from {:\ \,}
\makeatother

\newcolumntype{Y}{>{\centering\arraybackslash}X}

\newcommand{\ignore}[1]{}

\def\eqref#1{Eq.~(\ref{#1})}

\newcommand{\bx}{\mathbf{x}}

\newcommand{\by}{\mathbf{y}}

\newcommand{\bw}{\mathbf{w}}

\usepackage{flushend}
\newcommand{\etal}{{\em et al.}}

\newcommand{\papertitle}{Choosing Smartly: Adaptive Multimodal Fusion \\  for Object Detection in Changing Environments}

\title{\LARGE \bf
\papertitle
}

\author{Oier Mees$^\ast$\thanks{} \and Andreas Eitel$^\ast$  \and Wolfram Burgard
  \thanks{$^\ast$These authors contributed equally.}\thanks{
    All authors are with the Department of Computer Science, University of Freiburg, Germany. 
    \{meeso, eitel, burgard\}@informatik.uni-freiburg.de. This work has partly
been supported by the European Commission under ERC-AGPE7-267686-LIFENAV.}
}
\begin{document}
\maketitle
\thispagestyle{empty}
\pagestyle{empty}

\begin{abstract}
  Object detection is an essential task for autonomous robots
  operating in dynamic and changing environments. A robot should be
  able to detect objects in the presence of sensor noise that can be
  induced by changing lighting conditions for cameras and false depth
  readings for range sensors, especially RGB-D cameras.  To tackle
  these challenges, we propose a novel adaptive fusion approach for
  object detection that learns weighting the predictions of different
  sensor modalities in an online manner.  Our approach is based on a
  mixture of convolutional neural network (CNN) experts and
  incorporates multiple modalities including appearance, depth and
  motion.  We test our method in extensive robot experiments, in which
  we detect people in a combined indoor and outdoor scenario from
  RGB-D data, and we demonstrate that our method can adapt to harsh
  lighting changes and severe camera motion blur.
  Furthermore, we present a new RGB-D dataset for people detection in
  mixed in- and outdoor environments, recorded with a mobile robot. Code, pretrained models and dataset are available at  \url{http://adaptivefusion.cs.uni-freiburg.de}.
\end{abstract}

\IEEEpeerreviewmaketitle

%%%%%%%%%%%%%%%%%%%%%%%%%%%%%%%%%%%%%%%%%%%%%%%%%%%%%%%%%%%%%%%%%%%%%%%%%%%%%%%%%%%%%%%%%%%%%%%%%%%%%%%%%
%%%%%%%%%%%%%%%%%%%%%%%%%%%%%%%%%%%%%%%%%%%%%%%%%%%%%%%%%%%%%%%%%%%%%%%%%%%%%%%%%%%%%%%%%%%%%%%%%%%%%%%%%
\section{INTRODUCTION}

Most autonomous robots operating in complex environments are equipped
with different sensors to perceive their surroundings.  To make use of
the entire sensor information in the context of an object detection
task, the perception system needs to adaptively fuse the raw data of
the different sensor modalities.  This sensor fusion is challenging
for object detection because the sensor noise typically depends
substantially on the conditions of environment, which even might be
changing.  An example scenario for a changing environment would be a
robot equipped with an RGB-D sensor, that has to operate in both, a
dark indoor and a bright outdoor scenario, during different times of
the day, or under different weather conditions in the case of
autonomous cars. Our goal is to equip the perception system of robots
with the capability to autonomously adapt to the current conditions
without human intervention.  For example, in a dark indoor scenario,
one would expect the depth information from an RGB-D device to be more
reliable than the visual appearance.  Additionally, the depth stream
would not be very informative in a sunny outdoor scenario with objects
that are far away from the robot.  In this paper, we demonstrate that
such prior information can be learned from raw data and without any
hand-crafted features.  Therefore, how to best combine different
modalities for robust object detection is the main question we
tackle. We make the following contributions:
\begin{figure}[!ht]
  \centering
  \includegraphics[width=0.9\columnwidth]{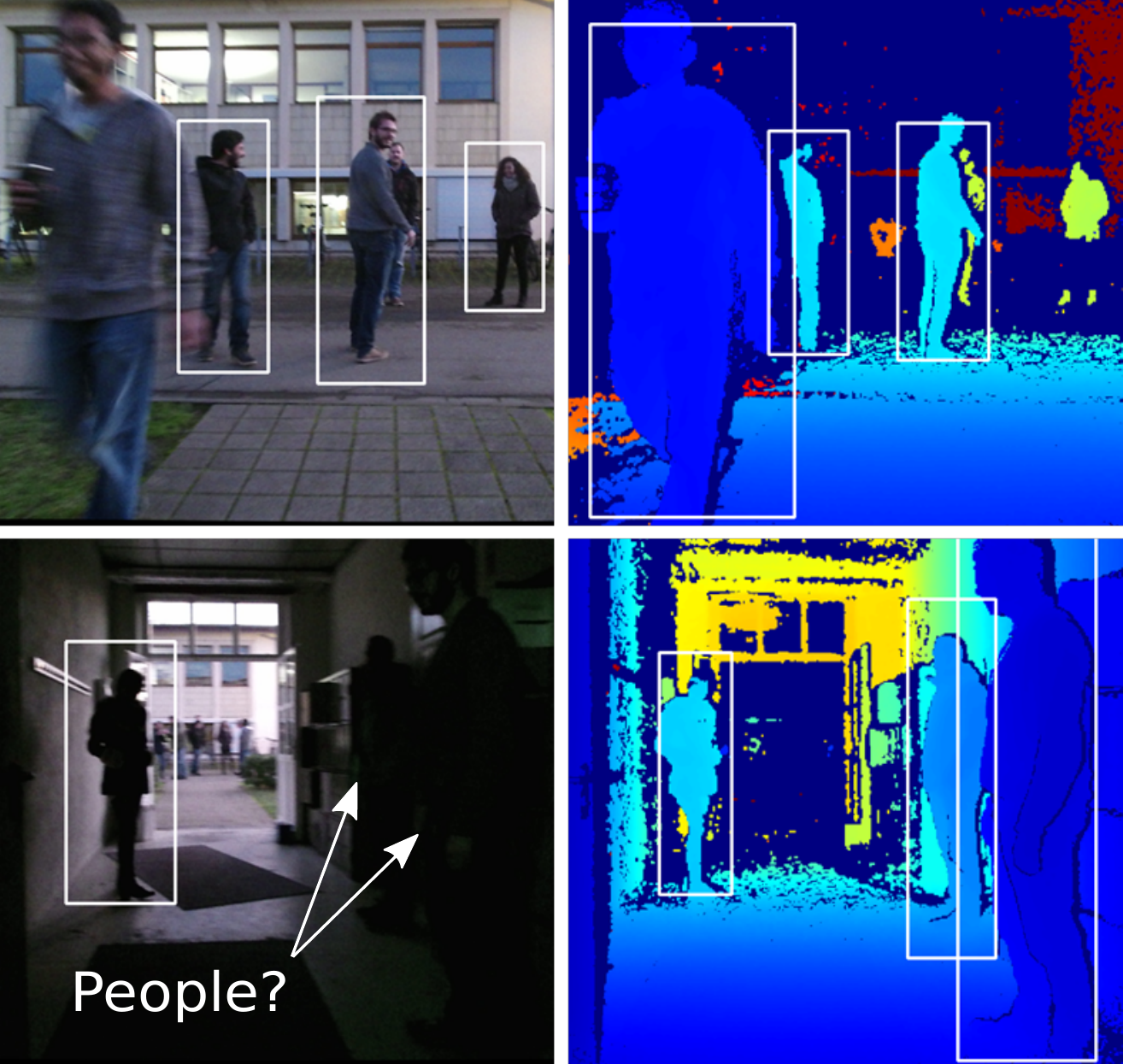}
  \caption{Vision-based detection fails in very dark environments and
    under motion blur, while it performs good in outdoor scenarios and
    for larger distances, where depth images are usually noisy. Our
    approach combines the best of both worlds in an adaptive fusion
    manner.\label{fig:front}}
\end{figure}
\begin{itemize}
\item We introduce a novel fusion scheme for object detection, based
  on a mixture of deep network experts.
\item We learn the adaptive fusion using a CNN, that is trained to
  weight the expert classifier outputs, based on high-level features
  extracted from the expert networks, without the use of prior
  information.
\item We evaluate our method in extensive real-world experiments and
  demonstrate that it is more robust in changing environments than
  purely vision-based or other multimodal fusion approaches.
\end{itemize}
Although our method is applicable for an arbitrary number of object
classes, in this work we report results for the two-class problem of
human detection in the RGB-D domain.  First, our experimental results
show an increased performance of our method compared to other fusion
approaches reported on the publicly available RGB-D People Unihall
dataset~\cite{spinello2011people}. Second, we evaluate our approach on
a more challenging detection scenario.  We recorded RGB-D sequences of
people from a mobile robot, captured under abrupt changes in lighting
conditions, both indoors and outdoors.  
In comparison to previously
recorded datasets for mobile robots, the sequences show the robot
moving through poorly illuminated indoor environments, followed by
very bright outdoor scenes within a short period of time.  Throughout
this paper, when talking about a changing or a dynamic environment, we
mean that the underlying conditions of the environment are changing.
Examples of underlying conditions are
\begin{itemize}
\item lighting changes,
\item out of range readings in the depth sensor, 
\item reflective materials that cause sensor noise in the depth
  channels of an RGB-D sensor and 
\item motion blur.
\end{itemize}

\begin{figure*}[t]
  \centering
  \includegraphics[width=0.86\linewidth]{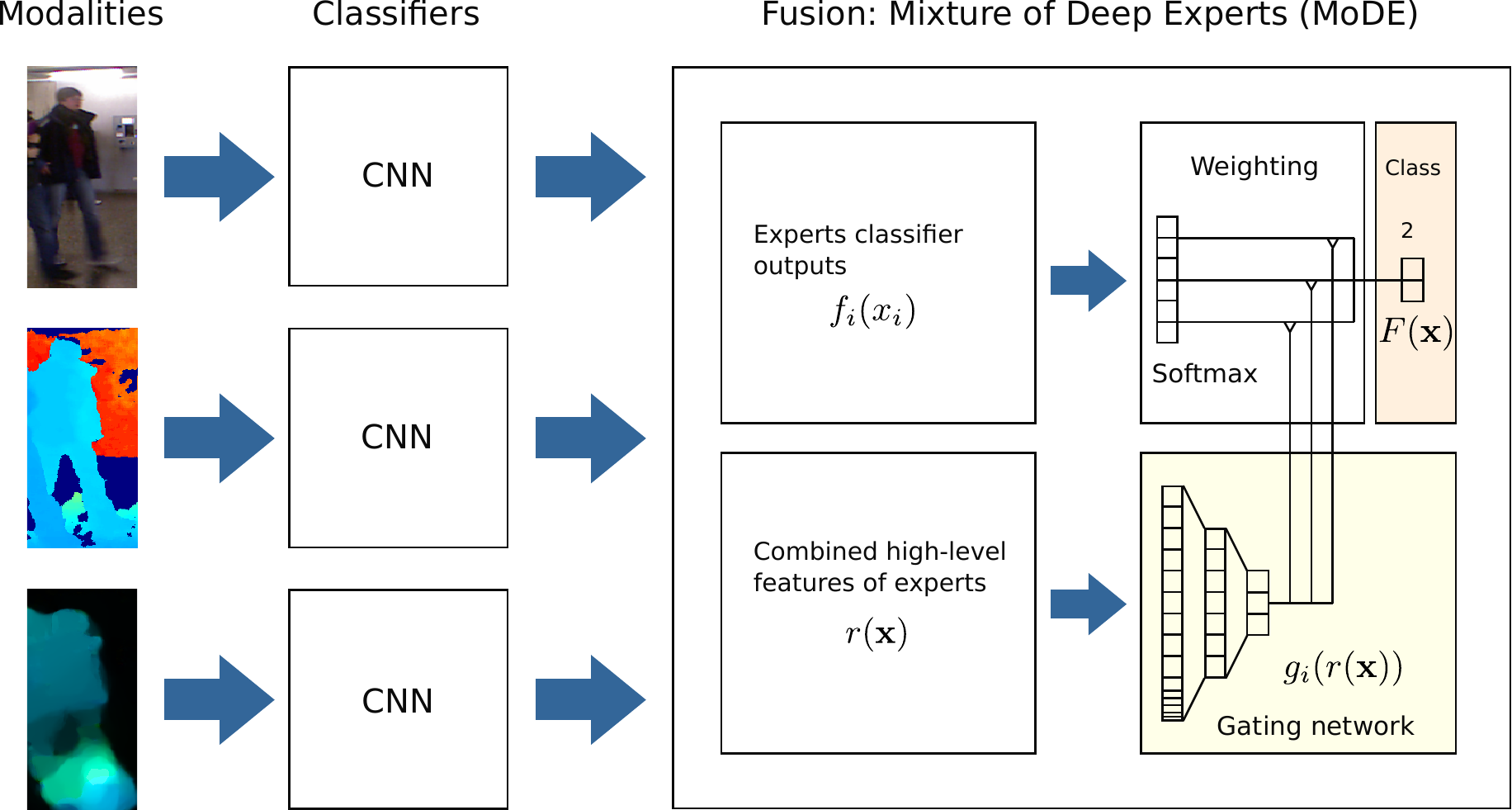}
  \caption{We propose a mixture of deep network experts for sensor
    fusion, which improves classification of several CNN experts
    trained on different input modalities through an additional gating
    network. The fusion is learned in a supervised way, where the
    gating network is trained with a feature representation of the
    expert CNNs. For each input sample, the gating network assigns an adaptive weighting value to each expert to produce the final classification output.  
    \label{fig:architecture2}}
\end{figure*}

\section{RELATED WORK}
Sensor fusion for robust object detection in changing environments is
at the core of many robotics applications. In the past, a large
fraction of work has focused on this task in the context of human and
also general object detection.  Yebes~\etal~\cite{yebes2015visual}
combined features from appearance and depth data in conjunction with a
DPM detector to recognize objects in road scenes.  The features are
merged in a channel fusion manner, by modelling a 3D-aware HOG
descriptor. In comparison to their method, our approach aims to
combine features at a later stage.  Enzweiler
\etal~\cite{enzweiler2011multilevel} introduced a mixture of experts
approach for people detection using three input modalities, namely
appearance, motion and depth.  In comparison to our approach the
weighting of the experts is constant and therefore not adaptive.
Premebida \etal~\cite{premebida2014pedestrian} trained a late fusion
SVM with manually designed features to perform detector fusion from
RGB and depth modalities. In our work, we do not use any prior
information to learn the weights for the fusion, although it would be
possible to use our method with manually designed fusion features.
Spinello \etal~\cite{spinello2012leveraging} proposed a hierarchical
mixture of experts approach, where the output of the individual
detectors is weighted based on missing information in the sensor
modalities.  Compared to their approach, our weighting function is
directly learned from a feature representation of the raw input data.
Very recently, late-fusion network architectures have shown to be very
successful for vision-tasks such as multimodal pedestrian
detection~\cite{schlosser, wagner2016multispectral}, RGB-D object
recognition~\cite{eitel15iros} and RGB-D object
detection~\cite{gupta2014learning}.  Therefore, we will test similar
late-fusion network architectures as baselines for comparison with our
approach.

Our work is also related to the area of multimodal people detection,
with emphasis on mobile platforms equipped with RGB-D sensors.
Jafari~\etal~\cite{hosseini2014real} proposed a multimodal approach
which combines a depth-based close distance upper body detector and an
appearance-based full body detector based on groundHOG
features. Detections of both modalities are fed into a
multi-hypothesis EKF-based tracking module.  In comparison to
filtering over time, our approach performs fusion on a per-frame
basis.  For further comparisons, we refer the reader to the work of
Spinello \etal~\cite{spinello2011people}, Munaro
\etal~\cite{munaro2014fast} and Linder
\etal~\cite{linderGirrbachIROSWS15}.

There has been a large body of research targeting pedestrian detection
in the vision community~\cite{Tian_2015_CVPR,ouyang2013joint}.  For a
concise overview of pedestrian detection we refer to a recent
discussion by Benenson \etal ~\cite{benenson2014ten}. They conclude
that improved detection performance has shown to be driven by the
design of better features, but also complemented by additional data,
such as image context and motion.  To demonstrate the quality of
features learned by a convolutional neural network, Hosang
\etal~\cite{hosang2015taking} reported improved results for pedestrian
detection, using pre-trained off-the-shelf CNNs.  More recently
Angelova \etal~\cite{AngelovaKV15} presented a convolutional network
architecture that processes larger areas of the RGB input image and
detects multiple pedestrians simultaneously, leading to a significant
speedup at test time. However, none of these approaches make use of
multiple modalities.

\section{Mixture of deep networks architecture}
Our detection approach is based on a mixture of deep network experts
(MoDE) that are fused in an additional network, which we will further
denote as gating network.  The overall architecture, which is
illustrated in Fig.~\ref{fig:architecture2}, is an extension of the
adaptive mixture of experts method~\cite{jacobs1991adaptive} that
differs in the modelling of the gating network. It takes as input the
feature representations extracted from a higher level in the hierarchy
of each expert network, instead of using the raw pixel input as
presented in the original method.  The gating network then decides,
based on its input, how to weight the outputs of each expert to
produce the final classifier output.  Let
$\mathcal{D} = \lbrace (\bx^{1}, \by^{1}), \dots, (\bx^{N}, \by^{N})
\rbrace$
be the training examples, where $\bx^k = (x_1, \dots , x_M)$ denotes a
sequence of matrices which describe $M$ different input modalities
along with a gating network $g$.  The output encoding the class labels
is defined as a $C$ dimensional vector $\by\in\mathbb{R}^C$ in one-hot
encoding.

\subsection{Fusion via mixture of experts}
%%%%%%%%%%%%%%%%%%%%%%%%%%%%%%%%%%%%%%%%%%%%%%%%%%%%%
We combine the classifier outputs $f_{i}(x_i)\in\mathbb{R}^C$ of
$i = 1, \dots , M$ experts -- one for each modality -- by gating
functions $g_{i}(r(\bx))\in[0,1]$ with
$\sum_{i=1}^{M} g_{i}(r(\bx)) = 1$.  Let $h$ denote a feature map
produced by the last pooling layer of each expert CNN. It can be
described as a three-dimensional array of size
$Nh \times Hh \times Wh$ (number of filters, height, width). The
resulting concatenated feature map $h_{\text{concat}}$ for M experts
is an array of size $(Nh\cdot M) \times Hh \times Wh$. The flattened
representation of $h_{\text{concat}}$ is denoted as $r(\bx)$ which is
a one-dimensional array of size
$1 \times (Nh\cdot M \cdot Hh\cdot Wh)$.

Therefore the gating functions depend on the input $\bx$ solely
through the representation $r({\bx})$, which significantly minimizes
the input dimension of the gating network.  Each $f_{i}(x_i)$ maps the
raw pixel input to the C outputs $c = 1 , \dots , C$ and the fused
classifier output can then be written as
\begin{equation}
F(\bx) = [F_c(\bx)]_{c=1}^C = \sum_{i=1}^{M} g_{i}(r(\bx)) f_{i}(x_{i}). 
\label{eq:MoE}
\end{equation}
This can be formulated as probability model so that
\begin{equation}
F_c(\bx) \simeq \sum_{i=1}^{M} p(e_{i} \mid r(\bx) )\cdot p(c \mid e_{i}, x_{i}) = p(c \mid \bx),
\end{equation}
where $p(e_{i} \mid r(\bx))$ denotes the probability of selecting an
expert $i$ and
$g_{i}(r(\bx)) = p(e_{i} \mid r(\bx)) = \text{softmax}(\xi_i)$
represents the contribution of a single expert to the final
probability over classes $p(c \mid \bx)$. The $i^{th}$ output of the
last inner product layer of the gating network is denoted as $\xi_i$
and the softmax function is defined as
$\text{softmax}(z) = \exp(z)/ \sum_{j} \exp(z_j)$.  We train the
combined gating network architecture using the cross-entropy loss that
is defined as:
\begin{equation}
L(\bw) = - \frac{1}{N} \sum_{k=1}^{N} {\by^k}^T \log F(\bx^k). 
\end{equation}\\
The final architecture of our model is depicted in
Fig.~\ref{fig:architecture2}. For the following detection tasks we
define $C=2$, representing the two classes background and human.

\subsection{Training a mixture of experts}
We train our model using a two stage approach, where in the first
stage we train the individual expert networks in an end-to-end manner
using stochastic gradient descent (SGD).  In this paper, we use
several expert architectures, a standard three-layer convolutional
neural network as a baseline and a more evolved deep network based on
the Google inception architecture~\cite{SzegedyInception}.  The
three-layer convolutional neural network is a small network designed
to solve the CIFAR-10 classification problem and will be further
depicted as CifarNet. It has also been proposed by Hosang
\etal~\cite{hosang2015taking} as a good baseline for people detection.
In Section~\ref{ssec:inout} we consider a downsampled Google
inception architecture, depicted as GoogLeNet-xxs.  For our first
experiment in Section~\ref{ssec:uni}, the CifarNet baseline already
outperforms previous reported approaches. In order to leave room for
improvement via sensor fusion, we report results with the CifarNet as
architecture for the single experts and the mixture of experts
network.  In our second experiment, we show that further improvements
over the CifarNet baseline can be achieved, by replacing the network
with an inception architecture.  For the GoogLeNet-xxs network, we
only use the layers of the original network up to the first softmax
classifier (``softmax0'').  For training the networks, we use Fast
R-CNN~\cite{girshickICCV15fastrcnn}, including the region of interest
pooling layer and the multi-task loss for bounding box regression.
The framework is wrapped around a modified version of the Caffe
library~\cite{jia2014caffe}.  All experts are trained using standard
parameters and in the first stage we apply dropout, as a regularizer,
only in the fully-connected layers.  Further, the networks of the RGB
modality are fine-tuned using the pre-trained models available from
the Caffe library~\cite{jia2014caffe}, whereas the experts for the
other domains are trained from scratch.  In the second training phase,
the gating network is trained on an additional validation set.  We
optimize the weights of the gating network using SGD and keep the
weights of the individual experts fixed, by setting the learning rate
of all layers to zero.  Although the weights of the experts are not
changed, we apply a modification to the expert layers for training the
gating network.  In all expert layers we now apply dropout, as a
special case of data augmentation, in order to improve the performance
of the gating network.  To generate region of interest proposals for
the networks, we implemented a dense multiscale sliding window
approach.  Throughout the paper we use different input modalities to
feedforward through the networks.  We mainly use a combination from
the following modalities: RGB, depth and optical flow (for
representing motion).  For optical flow computation we use the OpenCV
implementation of the method proposed by Brox
\etal~\cite{brox2004high}.

%%%%%%%%%%%%%%%%%%%%%%%%%%%%%%%%%%%%%%%%%%%%%%%%%%%%%%%%%%%%%%%%%%%%%%%%%%%%%%%%%%%%%%%%%%%%%%%%%%%%%   

%%%%%%%%%%%%%%%%%%%%%%%%%%%%%%%%%%%%%%%%%%%%%%%%%%%%%%%%%%%%%%%%%%%%%%%%%%%%%%%%%%%%%%%%%%%%%%%%%%%%%
%%%%%%%%%%%%%%%%%%%%%%%%%%%%%%%%%%%%%%%%%%%%%%%%%%%%%%%%%%%%%%%%%%%%%%%%%%%%%%%%%%%%%%%%%%%%%%%%%%%%%%
\section{Experiments}

\subsection{RGB-D People Unihall Dataset} 
\label{ssec:uni} 
We evaluate our approach on the publicly available RGB-D People
Unihall dataset provided by Spinello\textit{ et al.}
\cite{spinello2011people}. The dataset contains over~3,000 frames of
people passing by three vertically mounted Kinect cameras in a
university hallway.  As evaluation metrics, we compute average
precision (AP) and equal error rate (EER). We define the equal error
rate as the point in the precision-recall curve where precision and
recall values are equal. Adopting their no-reward-no-penalty policy,
we do not count true positives or false positives when a detection
matches an annotation of a partially occluded person.  For training,
we randomly select 700 frames from each of the three Kinect cameras
and extract positive samples from the annotated candidates that show
fully visible people.  For evaluation, we use the remaining~300 frames
from each of the three Kinects.  To set hyperparameters of the
learning procedure we evaluate all trained models on the training set,
choosing the best performing model for evaluation on the test set.
Moreover, the dataset does not provide a pre-defined train/test split.
Therefore we created five random train/test splits, to train and
evaluate our detector.  We obtained a standard deviation of $\pm$0.8
EER, showing a small influence of the splits chosen. For some
experiments we report results using both an intersection over union
(IoU) of~0.4 and~0.6, due to different evaluation metrics used in the
literature. Unless otherwise specified, we use an IoU of~0.6 for
evaluation, which is the area of the overlap between a predicted
bounding boxes and an annotated ground truth box.
%%%%%%%%%%%%%%%%%%%%%%%%%%%%%%%%%%%%%%%%%%%%%%%%%%%%%%%%%%%%%%%%%%%%%%%%%%%%%%%%%%%%%%%%%%%%%%%%%%%%%
\begin{table}[t]
    \centering
    \begin{tabular}{c c c c  c}
  Input & Method & IoU & AP/Recall  & EER \\
  \hline
  \hline
  Depth & CifarNet & 0.6 & 78.0/88.0  & 84.5\\
  RGB & CifarNet & 0.6 & 66.5/73.0 &  71.8\\
  OF & CifarNet & 0.6 & 52.9/58.2 & 58.0\\
  D-OF & CifarNet average & 0.6 & 69.3/34.2  & 78.6  \\ 
  D-OF & CifarNet late fusion & 0.6 & 78.6/86.0  & 83.7 \\
  D-OF &  CifarNet MoDE & 0.6 & 78.1/88.5  & 86.0  \\        
  RGB-D-OF &  CifarNet average & 0.6 & 77.0/83.8  & 79.2\\
  RGB-D-OF & CifarNet late fusion & 0.6 & 88.0/88.4  & 88.2\\
  RGB-D-OF & CifarNet MoDE  & 0.6 & \textbf{88.6}/\textbf{90.0}  & \textbf{89.3} \\         
    \end{tabular}
    \caption{Performance of different single and multimodal networks on the RGB-D People dataset.}
    \label{tab:results}
\end{table} 
\begin{table}[t]
  \centering
  \begin{tabular}{c c c c}
  Input & Method & IoU &EER \\
  \hline
  \hline
  Depth & HOD + Segm~\cite{Biswas_2013_7457} & 0.4 & 84.0   \\
  Depth & HOD + Sliding Window~\cite{spinello2011people} & 0.4 & 83.0   \\
  Depth & HOD~\cite{spinello2012leveraging} & 0.6 & 56.3   \\
  RGB-D &  Combo-HOD~\cite{spinello2011people} & 0.4 & 85.0\\
  RGB-D &   HGE*~\cite{spinello2012leveraging} & 0.6 & 87.4\\
  \hline
  Depth & Ours, CifarNet & 0.4 & \textbf{93.7} \\
  RGB-D-OF &  Ours, CifarNet MoDE & 0.6 & \textbf{89.3}\\
  \end{tabular}
  \caption{Comparison of our best performing models with other approaches reported on the RGB-D People dataset.}
  \label{tab:sota}
\end{table}    
\subsubsection{Single expert performance}
We compare all reported approaches on the RGB-D People dataset that
use HOD features against our baseline architecture (CifarNet) trained
on colorized depth data.  We further report performance for the three
single networks, trained on three input modalities, namely RGB, depth
and optical flow.  The depth images are pre-processed using the jet
colorization methodology presented in our own previous
work~\cite{eitel15iros}.  Table~\ref{tab:results} shows that the CNN
trained on depth data (Depth CifarNet) performs better in comparison
to the network trained on the RGB modality (RGB CifarNet), which
matches previous results on the dataset~\cite{spinello2011people}. We
hypothesize the reason for this to be due to illumination changes and
low contrast found on the RGB images.  Besides, these lead to a
noticeable amount of corruption when computing the optical flow
images.  Consequently, the optical flow CNN underperforms when
compared with the other experts, also because it is limited to detect
moving people only.  We show in Table~\ref{tab:sota}, that our Depth
CifarNet outperforms HOD based approaches.  When evaluating the
detection at IoU~0.4, our best network achieves an EER of~93.7\%,
which is a relative improvement of~9.7\% over the graph based
segmentation combination with HOD proposed by Choi \textit{et
  al.}~\cite{Biswas_2013_7457}.  We report an improvement of~$28.2\%$
over previous approaches when evaluating the same model at IoU~0.6.
%%%%%%%%%%%%%%%%%%%%%%%%%%%%%%%%%%%%%%%%%%%%%%%%%%%%%%%%%%%%%%%%%%%%%%%%%%%%%%%%%%%%%%%%%%%%%%%%%%%%%   

 \begin{figure*}[t]
  \centering
  \includegraphics[width=0.80\textwidth]{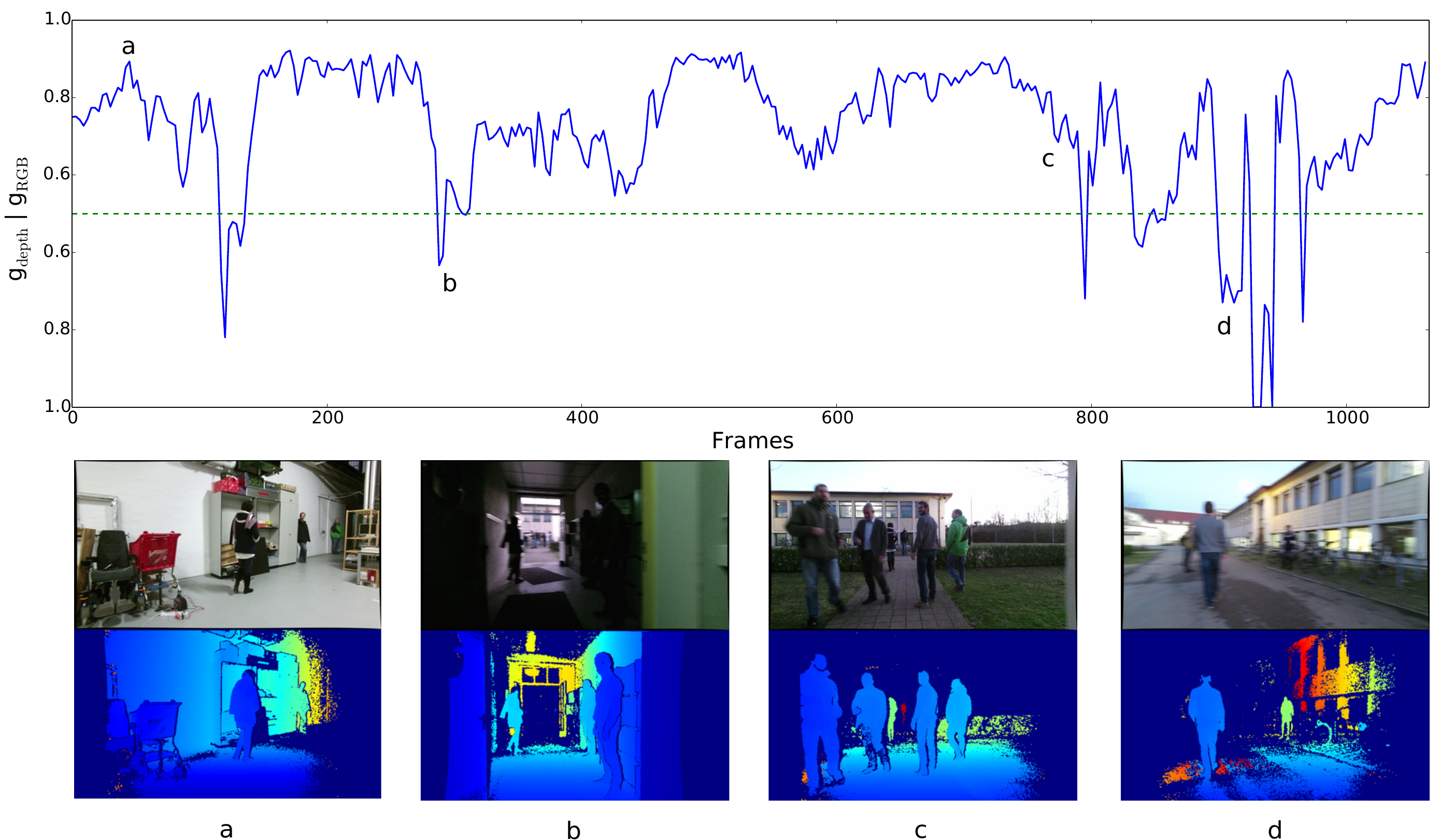}
  \caption{Timeline of the gating weights on the test sequence. The
    gating assignment weight switches between the RGB and depth
    modalities depending on the current environment (a, b, c, d). For
    example at frame b, the people are hard to detect in RGB, because
    the scene is dark. Accordingly, the assigned gating RGB weight is
    low $g_{\textrm{RGB}}=0.38$. The respective depth expert's gating
    weight is higher $g_{\textrm{depth}}=1-g_{\textrm{RGB}}=0.62$, due
    to people being better visible in the depth image.}
  \label{fig:weights_timeline}
\end{figure*}
\subsubsection{Comparison of fusion approaches} 
We compare detection performance of different fusion approaches when
combining 1) depth and motion (D-OF) and 2) depth, RGB and motion
(RGB-D-OF).  A naive way of fusing the individual detectors is
averaging the classifier outputs
$F(\bx) = \sum_{i=1}^{M} \frac{1}{M} f_{i}(\bx_{i})$.  As can be seen
in Table~\ref{tab:results}, this approach looses AP and EER points
with respect to other methods.  A more involved method is a late
fusion approach, where a fully-connected layer is trained on top of
the last pooling layers of all network streams.  Formally, we
concatenate the last layer feature responses $r(\bx)$ and feed them
through an additional two-layer fully-connected fusion stream
$f(r(\bx); \theta)$ with parameters $\theta$. The first
fully-connected layer has 64 outputs and is followed by a second layer
of output two, that ends in a softmax classifier. We follow the same
two stage training procedure described earlier.  The results given in
Table~\ref{tab:results} show, that the late fusion approach
substantially improves detector performance when combining all three
modalities. A combined depth and optical flow late fusion approach
underperforms and we conclude that it is not suitable for modalities
that highly differ in their feature space.  The architecture of the
gating network for the proposed MoDE consists of two
fully-connected layers with rectified linear units of size 64 and
three. The input of the gating network is the combined output of the
last pooling layer of each expert.  The proposed MoDE approach yields
best performance for all fusion scenarios. When combining all three
modalities (MoDE) we achieve an EER of 89.3$\%$.  In addition, we
report a relative improvement of 1.9\% EER over HGE's 87.4\%
EER~\cite{spinello2012leveraging}.
\subsection{InOutDoor RGB-D People Dataset} \label{ssec:inout} The
next experiment is conducted in a challenging people detection
scenario recorded from an Kinect v2 camera, that was mounted on a
mobile robot. Several abrupt and harsh lighting changes, combined with
severe motion blur provide challenges to both sensor modalities RGB
and depth, see Fig.~\ref{fig:front}. A particularity of the recorded
sequences is, that the robot is driving from indoor to outdoor
environments in a single take.  We recorded and annotated a total of
8605 RGB-D frames collected from the robot at a frame rate of
30hz. The camera was calibrated using the approach of
Wiedemeyer~\etal~\cite{iai_kinect2}.

The dataset is subdivided into four sequences, two recorded during
midday and the other two recorded at dusk.  The test sequence contains
1,066 frames, during which the robot drives from a dark indoor scene
into a relatively dark street scene, see
Fig.~\ref{fig:weights_timeline}.  
%The dataset will be available online
%at \url{http://adaptivefusion.cs.uni-freiburg.de}.
As evaluation metrics, we compute average precision (AP) and use an
IoU of~0.6 for evaluation of positive detections.

 \begin{figure*}[t]
  \centering
  \includegraphics[scale=0.42]{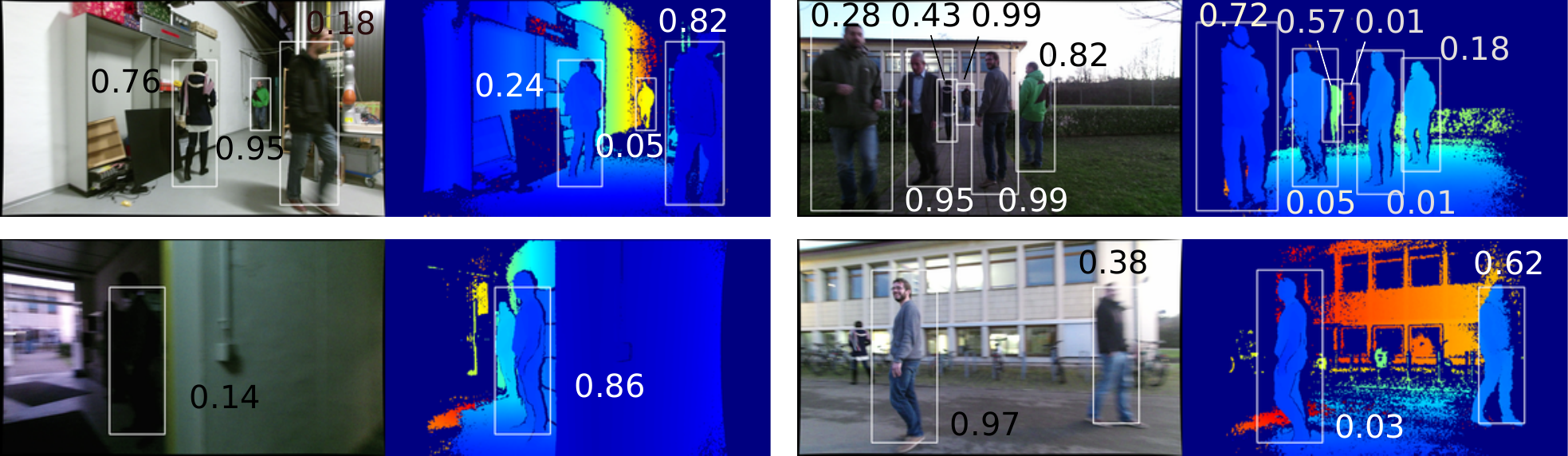}
  \caption{Qualitative results of people detection for the RGB-D based
    MoDE detector.  We show the assignment weights $g_i({r(\bx)})$
    learned by the gating network for each expert.}
  \label{fig:qualitative_results}
\end{figure*}

\subsubsection{Quantitative Results}
We report performance of several single modality networks, namely RGB
and depth and show improved results when fusing the modalities, see
Table~\ref{tab:InOutResults}.  Further, we compare the CifarNet
architecture with the more evolved GoogLeNet-xxs architecture.  All
single modality networks perform reasonably well on the test
set. Our depth-based GoogLetNet-xxs network outperforms the upper body detector of Jafari \etal~\cite{hosseini2014real}.
%However, the performance of the CifarNet is weaker than the
%cropped GoogLeNet.  
The gating network architecture for the
GoogLeNet-xxs consists of a convolution layer with output size~128
and kernel size~3, followed by two fully-connected layers of size~128
and~2. The combined outputs of the last convolutional layers of the
expert networks serve as input to the gating network.  Our novel
fusion approach improves the performance by 8.4\% AP in comparison to
the late fusion approach. The best model is a combination of the RGB
and depth GoogLeNet-xxs, fused with the gating network
(GoogLeNet-MoDE).  We also evaluated switching to the expert that the
gating network predicted, defining $g_{i}(r(\bx))\in\{0,1\}$ as in
\cite{jacobs1991adaptive}, instead of weighting the experts, but found
it to underperform. For further comparison, we trained a multi-channel
network with six channels as input in an end-to-end manner from
scratch. The network slightly underperforms in comparison to the late
fusion approach using the same GoogLeNet architecture.  The MoDE
method also outperforms a naive averaging of the classifier outputs.
\begin{table}[t]
    \centering
    \begin{tabular}{c c c}
  Input & Method & AP/Recall \\
  \hline
  \hline
  RGB & GoogLeNet-xxs  & 70.0/79.6 \\
  RGB & CifarNet & 55.3/62.9\\
  Depth & GoogLeNet-xxs  & 71.6/78.9\\
  Depth & Upper-body detector \cite{hosseini2014real} & 69.1/72.0 \\
  RGB-D & GoogLeNet average & 71.1/73.9 \\
  RGB-D & GoogLeNet channel fusion & 71.0/79.4 \\    
  RGB-D & GoogLeNet switching \cite{jacobs1991adaptive} & 71.0/73.9 \\ 
  RGB-D & GoogLeNet late fusion & 72.0/76.3 \\         
  RGB-D & GoogLeNet-MoDE  & \textbf{80.4}/\textbf{81.1} \\
  
    \end{tabular}
    \caption{Performance of single and multimodal networks on the InOutDoor RGB-D People dataset.}
    \label{tab:InOutResults}
\end{table}

\subsubsection{Qualitative results}
When analyzing the output detections obtained with the RGB-D based
MoDE detector over the full test sequence, we gain several interesting
insights.  Fig.~\ref{fig:weights_timeline} shows for each frame the
average of the gating weights for both experts.  The plot shows that
the average gating weights correlate reasonably well with the
respective environment. In the bright indoor scenario (sequence a),
the RGB modality is chosen more often, whereas in dark environments
(sequence b) the gating network tends to weight the depth network
output higher than the vision-based one. Outdoors (sequence c), it
mainly relies on RGB, especially for the pedestrian at far distance
who is almost not visible in the depth image, as can be seen in
Fig.~\ref{fig:qualitative_results}. We also found the depth network to
perform better for blurred images (sequence d).  Therefore, in the
final frames of the test sequence, the gating weights vary more in
between the frames. Here, the RGB outputs are mostly chosen, although
for abrupt camera motion the gating network switches with higher
frequency.  Fig.~\ref{fig:qualitative_results} shows the gating
weights for several exemplary frames, supporting our observation that
the gating network tends to switch for changing underlying conditions
such as lighting conditions, false depth readings at far ranges and
motion blur.
%%%%%%%%%%%%%%%%%%%%%%%%%%%%%%%%%%%%%%%%%%%%%%%%%%%%%%%%%%%%%%%%%%%%%%%%%%%%%%%%%%%%%%%%%%%%%%%%%%%%%   

%%%%%%%%%%%%%%%%%%%%%%%%%%%%%%%%%%%%%%%%%%%%%%%%%%%%%%%%%%%%%%%%%%%%%%%%%%%%%%%%%%%%%%%%%%%%%%%%%%%%%   

%%%%%%%%%%%%%%%%%%%%%%%%%%%%%%%%%%%%%%%%%%%%%%%%%%%%%%%%%%%%%%%%%%%%%%%%%%%%%%%%%%%%%%%%%%%%%%%%%%%%%%%%%
%%%%%%%%%%%%%%%%%%%%%%%%%%%%%%%%%%%%%%%%%%%%%%%%%%%%%%%%%%%%%%%%%%%%%%%%%%%%%%%%%%%%%%%%%%%%%%%%%%%%%%%%%
\section{Conclusion}
In this paper, we considered the problem of multimodal adaptive sensor
fusion for object detection in changing environments.  We proposed a
novel mixture of deep network experts approach that automatically
learns an adaptive strategy for weighting several domain-based
classifiers, from the raw input data. In extensive experiments, we
demonstrated that our multimodal method outperforms a vision-based
detection baseline and other fusion techniques. Moreover, we
show improved detection performance in a sequence recorded
from a mobile robot containing abrupt lighting changes and severe
motion blur.  Finally, our system outperforms previously reported
approaches for depth-based people detection on the publicly available
RGB-D People and our InOutDoor People dataset.


%%%%%%%%%%%%%%%%%%%%%%%%%%%%%%%%%%%%%%%%%%%%%%%%%%%%%%%%%%%%%%%%%%%%%%%%%%%%%%%%%%%%%%%%%%%%%%%%%%%%%%%%%
%%%%%%%%%%%%%%%%%%%%%%%%%%%%%%%%%%%%%%%%%%%%%%%%%%%%%%%%%%%%%%%%%%%%%%%%%%%%%%%%%%%%%%%%%%%%%%%%%%%%%%%%%
\section*{Acknowlegments}
We would like to thank Jost Tobias Springenberg, Luciano Spinello and Gian Diego Tipaldi
for helpful discussions. %This work has partly
%been supported by the European Commission under ERC-AGPE7-267686-LIFENAV.

%% Use plainnat to work nicely with natbib. 

\bibliographystyle{IEEEtrans}
\bibliography{references}
\addtolength{\textheight}{-12cm}   % This command serves to balance the column lengths

\end{document}